\pdfoutput=1
\documentclass[10pt,twocolumn,letterpaper]{article}

\usepackage{iccv}
\usepackage{times}
\usepackage{epsfig}
\usepackage{graphicx}
\usepackage{amsmath}
\usepackage{amssymb}

\usepackage{xcolor}
\usepackage{colortbl}

\usepackage[pagebackref=true,breaklinks=true,letterpaper=true,colorlinks,bookmarks=false]{hyperref}

\iccvfinalcopy 

\begin{document}

\title{Translating Visual Art into Music}

\author{Maximilian M\"uller-Eberstein\\
University of Amsterdam\\
{\tt\small edu@personads.me}
\and
Nanne van Noord\\
University of Amsterdam\\
{\tt\small n.j.e.vannoord@uva.nl}
}

\maketitle

\begin{abstract}
The Synesthetic Variational Autoencoder (SynVAE) introduced in this research is able to learn a consistent mapping between visual and auditive sensory modalities in the absence of paired datasets. A quantitative evaluation on MNIST as well as the Behance Artistic Media dataset (BAM) shows that SynVAE is capable of retaining sufficient information content during the translation while maintaining cross-modal latent space consistency. In a qualitative evaluation trial, human evaluators were furthermore able to match musical samples with the images which generated them with accuracies of up to 73\%.
\end{abstract}

\section{Introduction}
Art is experienced as a flow of information between an artist and an observer. Should the latter be visually impaired however, a barrier appears. One way to overcome this obstacle might be to translate visual art, such as paintings, from an inaccessible sensory modality into an accessible one, such as music.

Our research builds upon single-modality generative models for images \cite{vae, betavae} and music \cite{musicvae} as well as on multi-modal models which leverage corresponding audio-visual data in order to learn denser information representations \cite{pvae} or to make visual information more accessible \cite{chen2017deep}. Furthermore, generative models have been used to measure the expressiveness of image-based audio generation tools for the visually impaired \cite{ltti} and as such they offer a solid basis for our approach.

While previous approaches make use of paired audio-visual ground truth data, these are unavailable for visual art and music, making an unsupervised approach necessary. The main focus of this research will therefore be to learn a consistent cross-modal mapping in the absence of paired data such that similar images produce similar music. Evaluation metrics must be able to quantitatively reflect consistency and retained information content while a qualitative evaluation with humans must measure whether such consistencies are actively perceived.

To summarize, the main contributions of this research are as follows:

\begin{itemize}
	\item With the Synesthetic Variational Autoencoder (SynVAE), we introduce an unsupervised cross-modal architecture for translating data from one sensory modality into another consistently without the need for subjectively paired ground truth datasets (see Section \ref{sec:synvae}).
	\item In a series of experiments on generating music from the MNIST \cite{mnist} and Behance Artistic Media (BAM) \cite{bam} datasets, we compare a variety of mutual information metrics in order to establish a quantitative basis for evaluating such cross-modal models.
	\item In a qualitative study based on these quantitative metrics, we evaluate the human perception of the cross-modal translation consistency and lay out a framework for avoiding subjective biases within this process (see Sections \ref{sec:eval} and \ref{sec:experiments}).
\end{itemize}

\section{Methodology}
Leveraging the well established visual $\beta$-VAE architecture (VisVAE) \cite{betavae} and the auditive MusicVAE \cite{musicvae}, we are able to construct the Synesthetic VAE (SynVAE) (see Section \ref{sec:synvae}). Additionally, evaluating SynVAE with respect to cross-modal consistency requires a diverse set of methods which are outlined in Section \ref{sec:eval}.

\subsection{Synesthetic Variational Autoencoder}\label{sec:synvae}
\begin{figure*}
	\begin{center}
		\includegraphics[width=.7\textwidth]{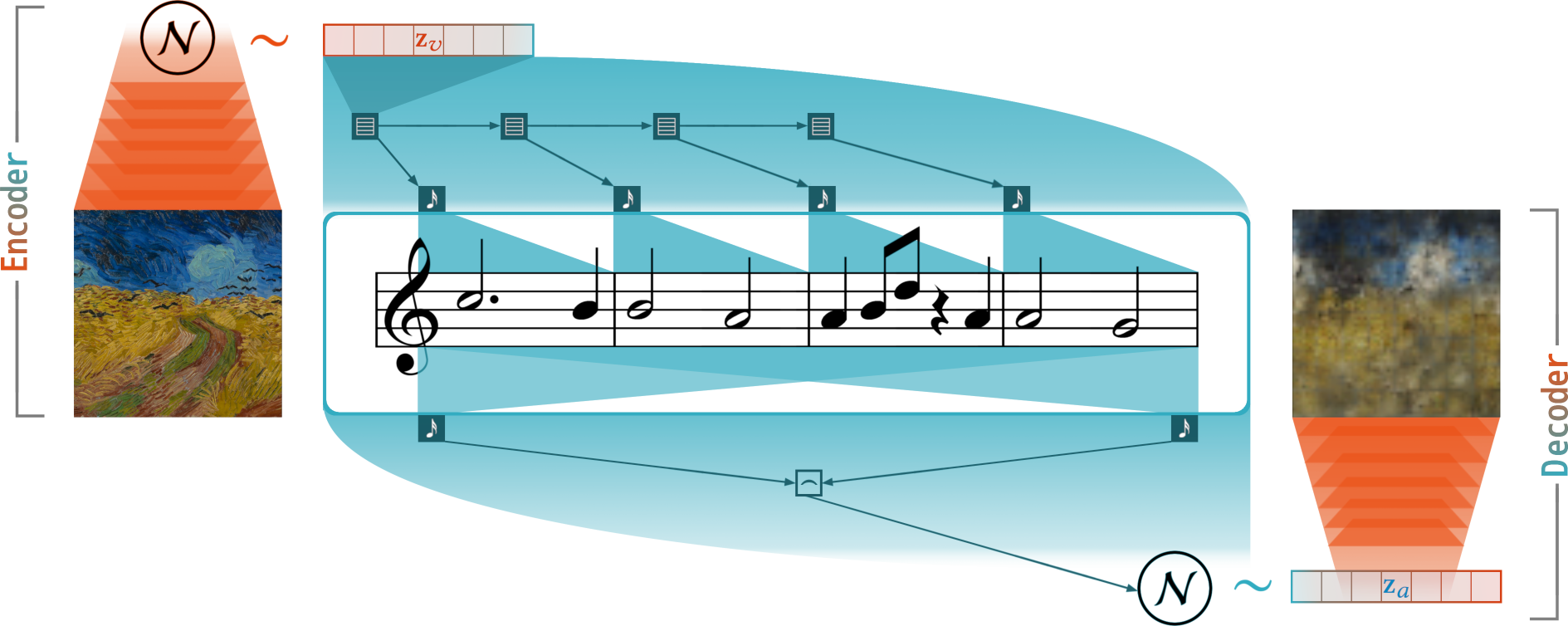}
	\end{center}
	\caption{Synesthetic VAE Architecture. An image is first encoded into a latent vector $\boldsymbol{z}_v$ by the VisVAE encoder, before being decoded into music by the MusicVAE decoder. During training, the music is subsequently re-encoded into $\boldsymbol{z}_a$ by the MusicVAE encoder and reconstructed into the output image by the VisVAE decoder.}
	\label{fig:synvae}
\end{figure*}

Translating information across the audio-visual modal boundary requires a synesthetic approach. By using multiple single-modality models which have unrestricted access to high quality data in their respective domains, we are able to remove the need for subjective correlations of images and music and are able to construct a fully unsupervised architecture which is outlined in Figure \ref{fig:synvae}.

Initially, the visual encoder $p_{\text{venc}}(\boldsymbol{z}_v|\boldsymbol{x})$ creates a 512 dimensional latent representation $\boldsymbol{z}_v$ from the original image $\boldsymbol{x}$. It provides the initial state of the pre-trained MusicVAE decoder $p_{\text{adec}}(\boldsymbol{a}|\boldsymbol{z}_v)$ which then produces a melodic output $\boldsymbol{a}$ using its hierarchical architecture. These two components make up the overall synesthetic encoder $p_{\text{senc}}(\boldsymbol{a}|\boldsymbol{x})$ which, during inference, can be used to perform the audio-visual translation.

In order to obtain a differentiable loss formulation, the audio output is subsequently re-encoded as $\boldsymbol{z}_a$ using the pre-trained, bidirectional MusicVAE encoder $p_{\text{aenc}}(\boldsymbol{z}_a|\boldsymbol{a})$ and then passed through VisVAE's decoder $p_{\text{vdec}}(\boldsymbol{x}'|\boldsymbol{z}_a)$ to produce an image reconstruction $\boldsymbol{x}'$. These two components make up the synesthetic decoder $p_{\text{sdec}}(\boldsymbol{x}'|\boldsymbol{a})$.

Both the visual and auditive latent vectors $\boldsymbol{z}_v$ and $\boldsymbol{z}_a$ in this model are sampled from multivariate Normal distributions that share the same regularizing prior distribution $\mathcal{N}(\boldsymbol{0}, I)$. This actively encourages the latent spaces to follow a similar and consistent shape across modalities when compared to an unregularized training procedure and is therefore at the core of our unsupervised approach.

Since this architecture requires an expressive musical latent space with high variability, the weights of the MusicVAE model remain fixed throughout the entire training process. Additionally, its pre-conditioning on the $\mathcal{N}(\boldsymbol{0}, I)$ prior alleviates us from the need to enforce a regularizing KL constraint on the frozen auditive components. Furthermore, audio reconstruction need not and cannot be measured due to the general absence of an audio-visual ground truth. The differentiable basis $\mathcal{L}_{\text{syn}}$ for the optimisation process therefore only consists of a KL constraint on the visual encoder, in addition to the difference of the synesthetic decoder's reconstruction with the original image:

\begin{equation}\label{eq:losssyn}
	\begin{aligned}
		\mathcal{L}_{\text{syn}} = &-\mathbb{E}_{\boldsymbol{a} \sim p_{\text{senc}}(\boldsymbol{a}|\boldsymbol{x})}[\ln p_{\text{sdec}}(\boldsymbol{x}|\boldsymbol{a})]\\
		&+ \beta \hspace{.1cm} \text{KL}(p_{\text{venc}}(\boldsymbol{z}_v|\boldsymbol{x}) \parallel p_{\text{prior}}(\boldsymbol{z}_v))
	\end{aligned}
\end{equation}
\vspace{.5em}

with $\beta$ controlling the balance between reconstruction quality and adherence to the canonical prior. In the synesthetic case, this carries additional importance since the actively trained visual components cannot stray too far from the prior without risking to land in undefined music space.

Extending this approach to further modalities is low friction. For instance, switching the auditive and visual components results in a SynVAE which encodes music into corresponding visuals. Regardless of the modality pair, as long as the encoders are regularized with matching prior distributions and the central generative component is capable of producing realistic results across its latent space, the cross-modal translation is likely to succeed. Since advances in single-modality models are being made constantly, this architecture could keep on improving in parallel to such generative models provided they offer consistent latent spaces.

\subsection{Evaluation}\label{sec:eval}

Apart from the quantitative metrics already present in the loss formulation $\mathcal{L}_{\text{syn}}$ (\ie MSE for reconstructions' information retention and KL divergence as a proxy for latent space consistency), labels of the visual datasets can be used to measure how well latent representations encode semantic similarity. By collecting the nearest neighbours of each encoded data point, it is possible to calculate the class precision at rank $n$. Since the higher inner-class visual variance of more complex datasets may not be well reflected however, we attempt to bridge this gap with reconstruction classification accuracy. Using the same architectures as for the visual encoders, but with a final softmax layer, classification networks are trained and tested on the images reconstructed by the autoencoders.

We are also strongly interested in the degree to which the visual and auditive latent spaces within the synesthetic model overlap. In absence of paired images and audio, we make use of Data-Efficient Mutual Information Neural Estimation (DEMINE) \cite{demine} in order to approximate a lower bound on the mutual information $I(\mathbf{Z}_v; \boldsymbol{Z}_a)$ of corresponding visual and auditive latent vectors.

Considering SynVAE's potential application in cultural settings such as art exhibitions, it is vital to evaluate whether humans can correlate similar images and audio in the same way as the model would predict. Since evaluating the entire corpus is not feasible and determining a sub-sample manually would introduce the curator's bias, we propose a more consistent and reproducible approach: By using VisVAE's latent space to sample images embedded closely around the centroids of their semantic classes, evaluators can be presented with a smaller subset of representative images and subsequently, their related audio. Through the accuracy with which they can identify correct audio-visual pairs, it is possible to determine whether the model's translations line up with human notions of similarity and intuitiveness.

\section{Experiments}\label{sec:experiments}

\begin{table}
	\begin{center}
	\begin{tabular}{lllllll}
	  	\hline
    	Model & MSE & KL & P10 & Acc & MI & Ql\\ 
    	\hline
		\textsc{M-Vis} & 16.43 & 24.38 & 0.31 & 0.99 & - & -\\
		\textsc{M-Syn} & 36.66 & 15.63 & 0.28 & 0.96 & 5.03 & 0.73\\
	    \arrayrulecolor{gray}
	    \hline
		\textsc{B-Vis} & 273.08 & 56.34 & 0.23 & 0.80 & - & -\\
		\textsc{B-Syn} & 455.16 & 27.89 & 0.25 & 0.77 & 5.16 & 0.71\\
		\arrayrulecolor{black}
	    \hline
	\end{tabular}
	\end{center}
	\caption{MSE, KL divergence, Precision@10 (P10) and classification accuracy (Acc) for VisVAE and SynVAE models on MNIST (M) and BAM (B) test sets. Mutual information lower-bound (MI) and evaluator accuracy on qualitative task (Ql) for SynVAEs.}
	\label{tbl:results}
\end{table}

\begin{figure*}
	\begin{center}
		\includegraphics[width=.95\textwidth]{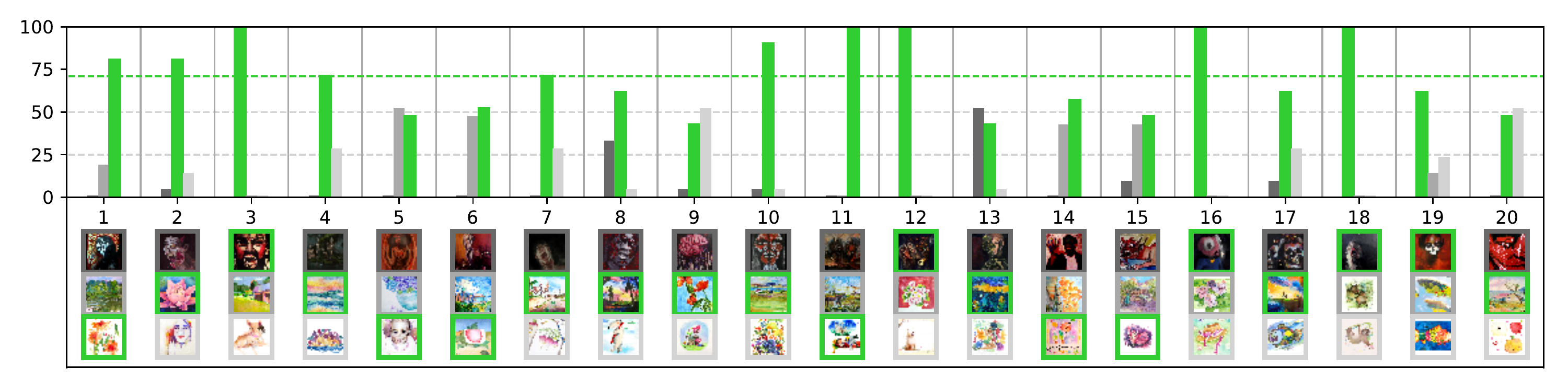}
	\end{center}
	\caption{Percentages of evaluator choices per class ("scary", "happy+peaceful", "happy") and task on the qualitative BAM evaluation. Correct options highlighted in green. Average accuracy marked at 71\%.}
	\label{fig:bam-qualres}
\end{figure*}

Our experiments involve the simple, but interpretable $28 \times 28 \times 1$ MNIST images \cite{mnist}, as well as the highly diverse BAM \cite{bam} of which we use $\sim$180k oil and watercolour paintings, downscaled and cropped to $64 \times 64 \times 3$. The latter's four emotion labels are of especial interest for the qualitative evaluation due to their abstract nature and reliance on shared human intuition. Additionally, we highly recommend listening to selected audio-visual examples from each dataset on \url{https://personads.me/x/synvae} (full source code is also available).

Both tasks share a similar pipeline: Visual $\beta$-VAEs are trained and evaluated before they are placed into SynVAE. It too is then trained, validated and tested on the respective splits of the same data using the methodology outlined in Section \ref{sec:synvae}. For each task, $\beta$-values in $[0.1, 2.0]$ were tested in a grid search pattern. Results for the $\beta=0.5$ MNIST and the $\beta=1.3$ BAM models are presented in Table \ref{tbl:results}.

From MSE and KL divergence, it is observable that passing visual information through music space results in a definite reduction in visual fidelity since VisVAEs have reconstruction errors which are close to half of that of their synesthetic counterparts. Regarding the adherence to the canonical prior however, the tables are turned: the SynVAEs have KL terms which are up to 61\% smaller. This stems from the strong regularizing effect of the fixed auditive components.

Measuring our main goal of audio-visual consistency was shown to be difficult using a single quantitative metric. Precision metrics based on the nearest neighbours of an embedded data point were indicative of consistency only if visual similarity was strongly correlated with labelled semantic similarity. For MNIST this was found to be the case since the monochrome nature of the dataset as well as the relatively low variance between images of the same class allow for MSE to be an appropriate surrogate for the detailed, low-level information content of an image.

For the more visually complex BAM with its exponentially larger amount of information as well as a higher in-class variance, it becomes more difficult for SynVAE to encode such lower-level details. Therefore not all visual details are retained when they are passed through SynVAE's auditive latent space and image reconstructions are limited to higher-level features such as overall colour and object placement. However, the emotion labels tend to share a larger degree of correlation with such features (\eg dark images being "scary") and allow for P@10 scores of around 0.25.

Reconstruction classification provides a more flexible way to measure cross-modal consistency. It relies on semantic labels as well, but has the benefit that it is indicative of whether same-class image data is consistently encoded and decoded. Across all experiments a drop in accuracy from VisVAE to SynVAE showed that while there is a certain degree of information loss when passing through latent spaces cross-modally, overall visual consistency of semantic classes is maintained. Reconstructed MNIST digits even bear a strong enough semblance to the original images to warrant a 0.96 accuracy even after passing through music space. SynVAE's less detailed BAM reconstructions also retain sufficient higher-level information in order for multi-class emotion to be identified with 0.77 accuracy.

Independent of labelled data, DEMINE measures a relatively high amount of mutual information between corresponding visual and auditive latent vectors ($\ln(10) \approx 2.3$ nats would for instance already be sufficient to encode MNIST's class information). This shows that SynVAE does indeed learn to embed information consistently across modalities and different types of visual data. Combining these three quantitative metrics, it becomes apparent that information content is indeed being translated across modalities consistently.

To assess whether human evaluators would share this notion of consistency, we conducted a classification study for MNIST and BAM with 11 and 21 participants respectively using the method described in Section \ref{sec:eval}. After being presented with 4 audio-visual example pairs for each of the 3 most distinct classes (as determined by VisVAE), evaluators were asked to identify which of 3 images was used to generate an audio across 20 trials.

For MNIST, evaluators achieve 0.73 accuracy ($\sigma=0.22$) and a consistent Fleiss' kappa of 0.48 when distinguishing between the digits "0", "1" and "4". In two tasks, evaluators unanimously made the correct connection. For the BAM classes "scary", "happy" and "happy+peaceful", accuracy is at a comparable 0.71 ($\sigma=0.13$) with a Fleiss' kappa of 0.46. The results in Figure \ref{fig:bam-qualres} further show how four "scary" images and one "happy" image were correctly identified by all evaluators. Considering that class-adherence for BAM emotions is not as easily identified as for MNIST digits, the level of agreement across tasks is remarkably high. This extends to mismatches such as in tasks 5, 6 and 13 as well, since they typically occur between the correct image and its visually closest alternative, but rarely for the most distinct incorrect option. This corroborates that the audio-visual translations of SynVAE are indeed consistent.

The high accuracy with which the evaluators were able to distinguish between the three most distinct classes of two very different datasets by ear alone, shows that low-level information, such as digits, can be conveyed audibly for simple data and high-level information, such as emotion perceived through colour and composition, can be conveyed for more complex data. This in addition to the quantitative results confirms that audio-visual consistency is not only theoretical, but also very perceivable.

\section{Conclusion}
As shown by our results, it can be concluded with high confidence that SynVAE is able to consistently translate a diverse range of images into the auditory domain of music through unsupervised learning mechanisms. The modular nature of SynVAE furthermore allows for this approach to be extended to any modality for which high quality single-modality datasets exist. We therefore hope that the methodology outlined in this research will provide a solid basis for evaluating unsupervised, cross-modal models, in addition to SynVAE itself enabling more intuitive and inclusive access to visual artworks across sensory boundaries.

\section*{Acknowledgements}
This research would not have been possible without additional valuable input from Marco Federici and Gjorgji Strezoski. Additional thanks go to the volunteer evaluators for their time.

{\small
\bibliographystyle{ieee}
\bibliography{references}
}

\end{document}